%% file: main.tex
\pdfoutput=1

\documentclass[11pt]{article}

\usepackage[]{acl}

\usepackage{times}
\usepackage{latexsym}
\usepackage{xspace}
\usepackage{verbatim}
\usepackage{graphicx}
\usepackage{cleveref}
\usepackage{hyperref}
\usepackage{colortbl}
\usepackage{xcolor}
\usepackage{colortbl}
\usepackage[T1]{fontenc}

\usepackage[utf8]{inputenc}

\usepackage{microtype}

\title{Exploring the Role of Task Transferability in Large-Scale Multi-Task Learning}

\author{
Vishakh Padmakumar \textsuperscript{1}\thanks{\hspace{3mm}Work done during summer internship at AWS AI.}~~~~~~Leonard Lausen\textsuperscript{2}~~~~~~Miguel Ballesteros\textsuperscript{3}~~~~~~Sheng Zha\textsuperscript{2}\\~~~~~~\bf{He He\textsuperscript{12}~~~~~~George Karypis}\textsuperscript{2}\\
\textsuperscript{1}New York University,
\textsuperscript{2}AWS AI, \textsuperscript{3}AWS AI Labs\\
{\tt vishakh@nyu.edu} \\
{\tt \{lausen, ballemig, zhasheng, hehea, gkarypis\}@amazon.com}\\
}

\input{macros}

\begin{document}
\maketitle
\begin{abstract}

Recent work has found that \mtl training with a large number of diverse 
tasks can uniformly improve downstream performance on unseen target tasks. 
In contrast, literature on task transferability has established that the choice of intermediate tasks can heavily affect downstream task performance. In this work, we 
aim to disentangle the effect of scale and relatedness of tasks in \mtl representation learning.
We find that, on average, increasing the scale 
of \mtl learning, in terms of the number of tasks, indeed results in better learned representations than smaller \mtl setups. 
However, if the target tasks are known ahead of time, then training on a smaller set of related  tasks  
is competitive to the large-scale \mtl training at a reduced computational cost.

\end{abstract}
\input{intro}
\input{approach}
\input{experiments}
\input{results}

\input{related}
\input{conclusion}
\section*{Acknowledgements}
We would like to thank Stefano Soatto for providing feedback on this draft as well as the anonymouos reviewers for their helpful comments.
\bibliography{all,custom}
\bibliographystyle{acl_natbib}

\appendix
\input{appendix}

\end{document}

%% file: macros.tex
\newcommand\todo[1]{\textcolor{red}{[TODO: #1]}}
\newcommand\vis[1]{\textcolor{green}{[Vis: #1]}}

\renewcommand\todo[1]{}
\renewcommand\vis[1]{}
\newcommand{\mtl}{multi-task\xspace}
\newcommand{\Mtl}{Multi-task\xspace}
\usepackage{array}
\newcommand{\PreserveBackslash}[1]{\let\temp=\\#1\let\\=\temp}
\newcolumntype{C}[1]{>{\PreserveBackslash\centering}p{#1}}
\newcolumntype{R}[1]{>{\PreserveBackslash\raggedleft}p{#1}}
\newcolumntype{L}[1]{>{\PreserveBackslash\raggedright}p{#1}}

\definecolor{slred}{RGB}{255, 189, 189}
\definecolor{cblue}{RGB}{135, 206, 250}
\definecolor{qagreen}{RGB}{205, 255, 204}

%% file: intro.tex
\section{Introduction}
\label{sec:intro}
\begin{figure*}[hbt!]
    \centering
    \includegraphics[width = \textwidth]{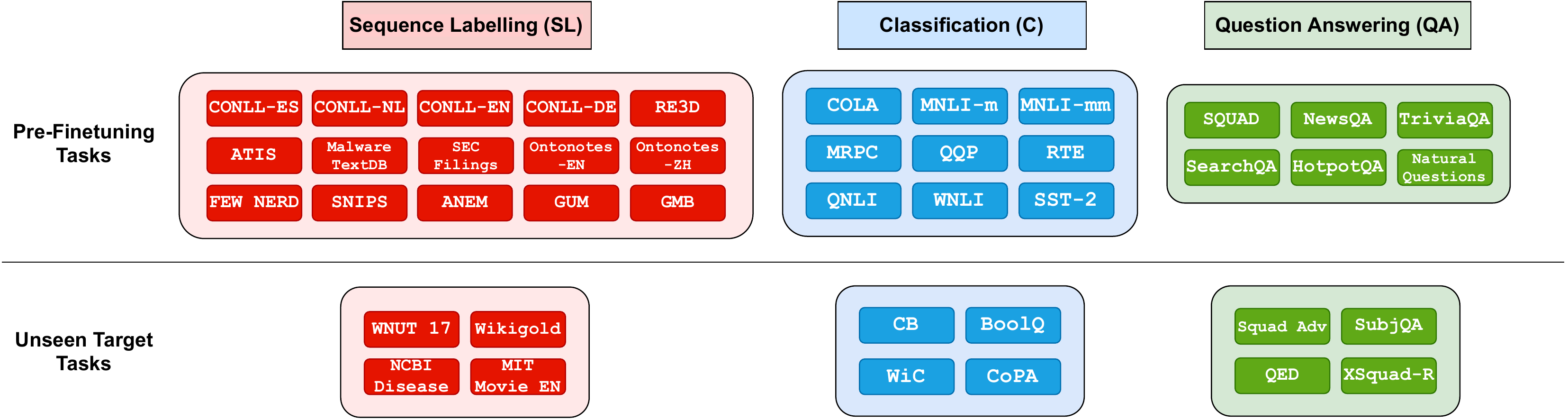}
    \caption{List of tasks grouped into sequence labelling (SL), question answering (QA) and classification (C).}
    \label{fig:tasks}
\end{figure*}

Following the wide success of unsupervised language model pre-training \cite{devlin2019bert, liu2019roberta, lewis2019bart}, recent work on transfer learning has shown that additional 
supervised \mtl training  
further improves performance 
on various downstream NLP tasks \cite{raffel2019exploring, khashabi2020unifiedqa, aghajanyan2021muppet}. There are two distinct ways in which the supervised data has been used: increasing the \emph{scale} of the \mtl step to incorporate more tasks \cite{khashabi2020unifiedqa, aghajanyan2021muppet} and developing task similarity metrics to incorporate tasks \emph{related} to the target task \cite{pruksachatkun2020intermediate, vu2020exploring}. 

\citet{aghajanyan2021muppet} show  
that a \mtl training step, or \emph{pre-finetuning} step, with a sufficiently large,  
diverse set of tasks is 
an effective task-agnostic second stage of model pre-training before finetuning on target tasks.  
In particular, they find that using a large number of tasks (e.g., roughly 15 tasks)
is crucial in achieving good downstream performance,
while pre-finetuning with fewer tasks 
causes a small performance drop. 
Meanwhile, work on task transferability \cite{vu2020exploring, pruksachatkun2020intermediate} has shown that the choice of 
individual intermediate  
tasks significantly affects 
downstream fine-tuning performance---predicting the transfer is challenging and there is high variance depending on the choice of intermediate task.  

This motivates us to ask the question if pre-finetuning on a small group of tasks related to the target tasks can obtain comparable performance to large-scale \mtl training. 
In this work, we present an empirical study to answer this question  
by extending the task transferability experiment 
to groups of tasks.  

We follow the two-step experimental pipeline from \citet{aghajanyan2021muppet}, 
where a pre-trained model is first pre-finetuned on a set of tasks and then separately finetuned on various target tasks on which we report performance. 
In addition, we group our set of 29 pre-finetuning 
tasks based on task format  
into 3 groups---classification tasks, sequence labelling tasks and extractive question answering tasks (\Cref{fig:tasks}). 
We perform model pre-finetuning on every combination of these 
task groups and report performance on target tasks that belong to each group. 
This allows us to systematically study how the size  
of the \mtl step 
and the choice of pre-finetuning tasks affects downstream task performance.

We observe that, on average, large-scale \mtl pre-finetuning results in improved 
performance on
downstream target tasks. We also see that  
a model trained on 
related\footnote{In the rest of this work, when we say that two tasks are related it means that they belong to the same task group} pre-finetuning tasks  
obtains comparable downstream task performance  
to the large-scale model, at a reduced computational cost\footnote{We say that a pre-finetuning run is of a cheaper than another in terms of computational cost when it involves \mtl training on fewer examples (\Cref{sec:cost})}
, but 
pre-finetuning on an unrelated grouping can result in a severe decline 
in 
performance. 

Our findings show the interplay between \mtl scaling and task selection---when the target tasks are unknown then \mtl scaling is an effective, if expensive, intermediate step 
but if the goal is to improve performance on a specific target task set then \mtl training on a smaller set of related tasks is an effective cheaper alternative.
These results also hint at the need to better study modeling techniques that mitigate negative transfer between tasks in large-scale \mtl training. 

%% file: approach.tex
\section{\Mtl Setup}
\label{sec:approach}
\begin{figure}
    \centering
    \includegraphics[width=0.45\textwidth]{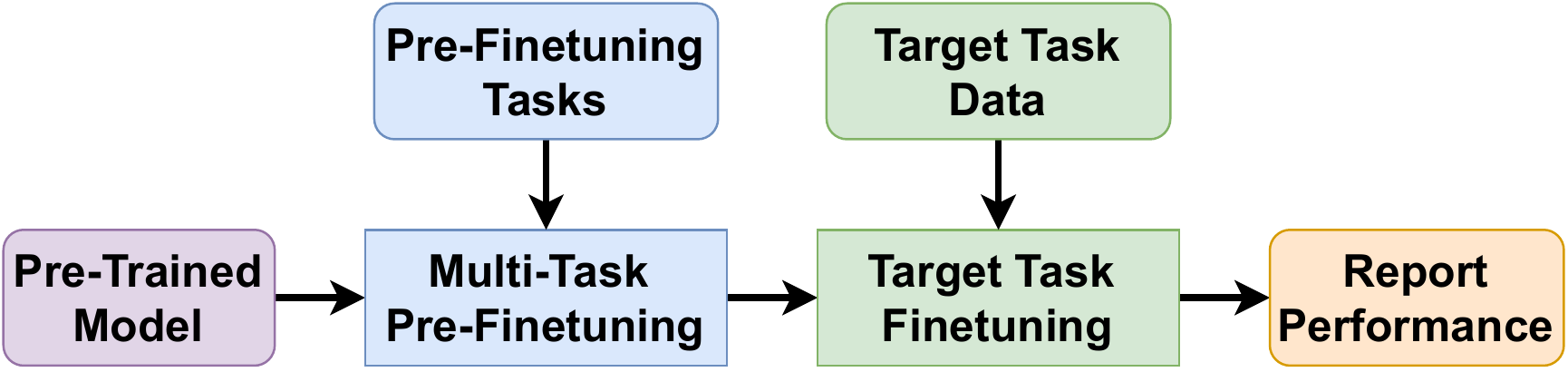}
    \caption{High level workflow of the \mtl setup.}
    \label{fig:workflow}
\end{figure}


Our \mtl experiments follow the two step approach from \citet{aghajanyan2021muppet}, shown in \Cref{fig:workflow}. A pre-trained model is first pre-finetuned on a set of tasks to obtain a single shared encoder. This model is then finetuned separately on various target tasks 
on which we report performance. 

 

\paragraph{Grouping of Tasks}  \Cref{fig:tasks} shows our list of 29 pre-finetuning tasks 
and 12 unseen target tasks (listed in \Cref{sec:datasets}). 
Informed by prior work \cite{vu2020exploring, ye2021crossfit, sanh2021multitask}, we divide these datasets into three groups based on task format
---sequence labelling (SL), extractive question answering (QA) and classification (C). As noted in \citet{sanh2021multitask}, grouping of NLP tasks is an imperfect heuristic. Prior work \cite{achille2019task2vec, vu2020exploring} formalizes the notion of task similarity using learned task embeddings so an alternate formulation would be to divide the tasks into groups based on these learned embeddings. In this work we focus on a simple, intuitive grouping based on the task format, or output space, of the task. 

\paragraph{Research Question} We aim to study how the choice of pre-finetuning tasks and the size of the \mtl step, in terms of number of pre-finetuning tasks, affects target task performance. 
In order to do so, 
we compare pre-finetuning runs on all combinations of task groups, reporting performance on target tasks from each group.

%% file: experiments.tex
\section{Experiments}

\subsection{Model Details}
We use the pre-trained XLM-Roberta model 
\cite{conneau2020unsupervised} 
for all of our experiments. During  pre-finetuning, we learn a shared encoder for all tasks and a task-specific head  
for each pre-finetuning task. For downstream finetuning, we randomly initialize a new head for each target task. We use the Huggingface \cite{wolf-etal-2020-transformers} XLM-Roberta base pre-trained model. The various task specific heads are linear classifiers on the encoder output per the Huggingface implementation. More model details are provided in \Cref{sec:app_model}.

\subsection{Training Details}
During pre-finetuning, to ensure that the model does not overfit to any particular task, 
we follow the sampling approach from \citet{aghajanyan2021muppet} of ensuring that each batch consists of examples from multiple tasks while maintaining the empirical distribution  
of examples in the datasets. 
For our study, the loss function for all the different pre-finetuning tasks is cross entropy loss. 
\citet{aghajanyan2021muppet} recommends scaling the losses from each task-specific head based on the size of the label space to ensure more stable training.  Our preliminary experiments showed better results without loss scaling so we follow the same for all pre-finetuning runs. 

We run pre-finetuning on the powerset of the set of task groups from \Cref{fig:tasks} and for each pre-finetuning run we report performance on all the target tasks. Hyperparameters are kept uniform across pre-finetuning runs 
and we train with batch size 128 and an early stopping criteria based on validation loss.  
Once pre-finetuning is completed, we finetune the model on each target task and report the average validation set performance\footnote{For all sequence labelling target tasks, we report F1 score. Among classification tasks, we report F1 for CB and accuracy for the other tasks. For extractive question answering, we report F1. These choices was made based on the standard evaluation for the task from the Huggingface metrics module.} across 5 
random seeds along with the associated standard deviation. 
More details on pre-finetuning and target task finetuning are included in \Cref{sec:app_pft}.

\paragraph{Notation} 
We refer to the pre-finetuning runs 
with their initials---Only-QA means that only 
question answering tasks were used in pre-finetuning the model and QA+C means that all question-answering and classification tasks were used and so on. We have 3 task groups and run experiments on all 7 possible combinations---Only-SL, Only-QA, Only-C, QA+C, QA+SL, SL+C and SL+QA+C. The baseline is a pre-trained model directly finetuned on the target tasks.

%% file: results.tex
\section{Results}

\setlength\tabcolsep{5pt}
\begin{table*}[h!]
    \centering
    \tiny
    \begin{tabular}{ c  c c c c c c c c }
        \hline
        \textbf{Dataset} & \textbf{Baseline} & \textbf{Only-SL} & \textbf{Only-C} & \textbf{Only-QA} & \textbf{SL+C} & \textbf{SL+QA} & \textbf{QA+C} & \textbf{SL+C+QA}\\
        \hline
        \cellcolor{slred}\textbf{WNUT17} & $63.239_{ \; 0.33}$ & $63.124_{\; 0.65}$ & $43.461_{\; 1.85}$ & $60.623_{\; 1.85}$  & $59.729_{\; 1.25}$ & $64.261_{\; 0.79}$ & $57.560_{\; 3.62}$ & $\mathbf{64.359_{\; 0.57}}$\\
        \cellcolor{slred}\textbf{Wikigold} & $80.409_{\; 1.43}$ & $\mathbf{83.449_{\; 0.16}}$ & $72.174_{\;	8.05}$ & $78.173_{\;	1.04}$ & $81.403_{\; 0.51}$ & $83.114_{\; 0.57}$ & $76.023_{\; 1.21}$ & $81.635_{\; 0.99}$\\
        \cellcolor{slred}\textbf{NCBI Disease} &	$86.849_{\; 0.46}$	&	$87.069_{\; 0.55}$ & $85.412_{\; 0.59}$	& $86.456_{\; 0.86}$ &	$87.037_{\; 0.43}$	&	$\mathbf{87.458_{\; 0.32}}$ & $86.642_{\; 0.70}$	& $87.378_{\; 0.73}$\\
        \cellcolor{slred}\textbf{MIT Movie} & $\mathbf{90.042_{\; 	0.53}}$ & $89.736_{\; 0.20}$	& $88.436_{\; 0.68}$ & $89.522_{\; 0.24}$ & $89.645_{\; 	0.27}$ & $89.829_{\; 0.24}$	& $89.287_{\; 0.20}$ & $89.631_{\; 0.21}$  \\ 
        \hline
        \cellcolor{cblue}\textbf{BoolQ} & $73.804_{\; 1.15}$ & $70.761_{\; 0.72}$ & $78.313_{\; 0.48}$ & $77.140_{\; 0.71}$ & $78.457_{\; 0.50}$ & $76.427_{\; 0.81}$ & $\mathbf{80.147_{\; 0.27}}$ & $79.515_{\; 0.51}$\\	
        \cellcolor{cblue}\textbf{CB} & $86.333_{\; 4.20}$ & $82.121_{\; 5.17}$ & $87.161_{\; 1.07}$ & $68.930_{\; 10.65}$ & $84.336_{\; 4.20}$ & $85.003_{\; 3.92}$ & $80.638_{\; 10.67}$ & $\mathbf{89.565_{\; 2.16}}$\\
        \cellcolor{cblue}\textbf{Copa} & $ 52.166_{\; 4.09}$ & $53.500_{\; 2.56}$ & $55.333_{\; 3.39}$ & $51.833_{\; 3.76}$ &  $53.666_{\; 3.20}$ & $53.666_{\; 2.49}$ & $54.333_{\; 1.59}$ & $\mathbf{57.333_{\; 1.24}}$\\
        \cellcolor{cblue}\textbf{WiC} & $60.136_{\; 7.45}$ & $63.244_{\; 1.01}$ & $64.812_{\; 0.51}$ & $ 54.937_{\; 6.56}$ & $65.229_{\; 1.45}$ & $65.177_{\; 0.61}$ & $65.151_{\; 1.35}$ & $\mathbf{65.674_{\; 1.05}}$\\
        \hline
        \cellcolor{qagreen}\textbf{Squad Adv} & $47.064_{\; 2.42}$ & $22.341_{\; 2.38}$ & $55.067_{\; 0.83}$ & $\mathbf{82.798_{\;	0.51}}$ & $52.035_{\; 1.05}$ & $81.558_{\; 	0.80}$ & $82.778_{\; 0.61}$ & $81.834_{\; 0.88}$\\
        \cellcolor{qagreen}\textbf{SubjQA} & $60.917_{\; 0.74}$	& $58.577_{\; 0.47}$ & $61.741_{\; 0.29}$ & $61.889_{\; 0.31}$ & $60.718_{\; 0.10}$	& $62.367_{\; 0.22}$ & $62.578_{\; 0.47}$ & $\mathbf{62.886_{\; 0.56}}$ \\
        \cellcolor{qagreen}\textbf{QED} & $37.779_{\; 5.49}$ & $36.544_{\; 1.16}$ & $51.604_{\; 1.34}$ & $74.643_{\; 0.44}$ & $46.533_{\; 1.21}$ & $\mathbf{77.196_{\; 0.51}}$ & $73.910_{\; 1.42}$ & $76.312_{\; 0.82}$\\
        \cellcolor{qagreen}\textbf{XQuad-R} & $63.522_{\; 3.66}$ & $50.398_{\; 1.83}$ & $64.538_{\; 0.54}$ & $\mathbf{80.744_{\; 0.50}}$ & $63.654_{\; 0.26}$ & $79.624_{\; 0.76}$ & $78.249_{\; 1.04}$ & $80.163_{\; 0.51}$ \\
        \hline
        \textbf{Average} & $68.312$ & $64.475$ & $68.298$ & $72.385$ & $69.147$ & $75.564$ & $74.160$ & $\mathbf{76.483}$ \\
        \textbf{Average Std. Dev.} & $2.662$ & $1.406$ & $1.637$ & $2.285$ & $1.202$ & $1.003$ & $1.929$ & $0.852$ \\
        \hline
    \end{tabular}
    \caption{
    Results on all the target tasks (rows) for all the pre-finetuning schemes (columns). Each cell value is an average on 5 runs with different seeds and the corresponding subscript is the standard deviation over these values. We also report the average over all tasks and the average of the standard deviation values in separate rows for analysis. We observe the effect of scale by seeing that the SL+QA+C setup has the best average performance across all tasks. We see that \mtl training results in reduced variability across multiple runs. 
    }
    \label{tab:unseen_tasks}
\end{table*}

\begin{table*}[h!]
\centering
\begin{tabular}{c c c c c c c c c}
\hline
 & \textbf{Baseline} & \textbf{Only-SL} & \textbf{Only-C} & \textbf{Only-QA} & \textbf{SL+C} & \textbf{SL+QA} & \textbf{QA+C} & \textbf{SL+C+QA}  \\
\hline
\cellcolor{slred}\textbf{Unseen SL} & $80.134$      & $\mathbf{80.844}$  & $72.370$  & $78.693$  & $79.453$ & $80.165$  & $77.378$ & $80.750$        \\
\cellcolor{cblue}\textbf{Unseen C}             & $68.109$      & $67.406$  & $71.404$ & $63.21$   & $70.422$ & $70.068$  & $70.067$ & $\mathbf{73.021}$ \\
\cellcolor{qagreen}\textbf{Unseen QA} & $56.692$ & $45.174$  & $61.120$  & $75.252$  & $57.568$ & $75.460$   & $75.035$ & $\mathbf{75.678}$   \\  
\hline
\textbf{Average} & $68.312$ & $64.475$ & $68.298$ & $72.385$ & $69.147$ & $75.564$ & $74.160$ & $\mathbf{76.483}$ \\
        \hline
\end{tabular}
\caption{
Results on all unseen tasks aggregated by task format (rows) for each pre-finetuning setup (columns). Each value in this table is an average of the 4 unseen tasks of that particular task format from \Cref{tab:unseen_tasks}. We see the effect of transferability where the Only-SL and Only-QA setups are competitive with SL+QA+C on unseen SL and QA tasks but suffer significantly on mismatched task groups. 
}
\label{tab:task_groups}
\end{table*}

\Cref{tab:unseen_tasks} shows the results of each pre-finetuning run on all the target tasks.  
\Cref{tab:task_groups} contains the results aggregated by task group. 


\paragraph{A large-scale target task-agnostic \Mtl step improves downstream performance} From \Cref{tab:unseen_tasks}, we see that the \mtl setup containing all 29 pre-finetuning tasks (SL+QA+C) has the best average performance over all the target tasks as well as the best score in 5 individual tasks. 
This is 
consistent with observations reported in \citet{aghajanyan2021muppet} that increasing the scale of the \mtl step results in better downstream performance on average across all tasks. 
Our results show 
the same trend on a different set of tasks with a smaller batch size.
We also show that this observation holds in a standard \mtl training regime, without the optimization tricks used \citet{aghajanyan2021muppet}, namely loss scaling and regularized finetuning \cite{aghajanyan2020better}.

\paragraph{Related tasks transfer better} To identify the role of transferability
, we aggregate the results on target tasks based on task groups in  \Cref{tab:task_groups}. 
Each row in \Cref{tab:task_groups} is the average of all the unseen tasks within that group 
from \Cref{tab:unseen_tasks}.  From this, we see that the Only-SL and Only-QA setups are on-par with SL+QA+C on unseen SL and QA tasks respectively indicating that pre-finetuning on a smaller set of related tasks obtains comparable performance to the large-scale \mtl model. 
  We also see that selecting a mismatched set of pre-finetuning tasks significantly hurts downstream task performance. 
From \Cref{tab:task_groups}, we see a drop of 9.6\% compared to the baseline on SL tasks with Only-C pre-finetuning and 
a 20.3\% drop on QA tasks with Only-SL.
These results extend 
those observed in \citet{pruksachatkun2020intermediate} and \citet{vu2020exploring} 
to transferability across task groups. 
With appropriate task selection, we can obtain comparable performance to the large-scale \mtl model at a reduced computational cost. 
The reduction in computational cost is mainly due to the change in number of pre-finetuning examples. We provide the comparison over various runs in \Cref{sec:cost}.
\citet{aghajanyan2021muppet} reported that \mtl learning is detrimental to target task performance at a smaller scale ($< 15$ tasks)
. In our study,  
we see that pre-finetuning on a single group of related tasks always outperforms the results from the mismatched pairwise setup---Only-QA outperforms SL+C on unseen QA tasks, Only-SL outperforms QA+C on SL tasks and Only-C outperforms SL+QA on C tasks.
Hence we conclude that at smaller scales, the particular pre-finetuning tasks selected significantly impacts downstream task performance, linking back to transferability literature.  

\paragraph{Tasks interact differently, so selecting an optimal subset is hard}

When we look at pre-finetuning runs on pairs of task groups taken together, 
we see that the SL+C and QA+C pre-finetuning setups perform 
worse on QA and SL tasks than even the baseline model but 
the SL+QA setup is competitive with  
the best SL+QA+C setup across all unseen tasks. 
This shows that selecting an optimal combination of tasks can be challenging based on task group heuristics. \citet{aribandi2021ext5} also observed a similar result that, at larger scales, a random subset of tasks often outperforms subsets selected using simple heuristics.  

\paragraph{\Mtl training reduces the variability of downstream task performance} In \Cref{tab:unseen_tasks}, we also report the standard deviation in target task performance across 5 random restarts. We see a trend that large-scale \mtl pre-finetuning reduces the variability across runs on all tasks---the SL+QA+C setup has the lowest average and the pairwise setups 
average lower variation than the single task group setups. \citet{phang2018sentence} also reported similar findings that \mtl training reduces variability in performance across random restarts. 
Additionally we observe that the Only-SL, Only-C and Only-QA setups have lower variability on unseen tasks of the same group than other groups, indicating that the downstream performance is more reliable on tasks within the same group.   
We discuss some limitations of our setup in \Cref{sec:limitations}

%% file: related.tex
\section{Related Work}

\paragraph{Large-Scale \Mtl Learning} Post the wide success of unsupervised language model pre-training, \citet{phang2018sentence} showed that intermediate task training on large datasets results in performance improvements on the GLUE benchmark. 
\citet{liu2019multi} showed an improvement over standard pre-training on multiple NLP 
benchmarks in the \mtl setting. T5 \cite{raffel2019exploring} framed various NLP tasks in a text-to-text format and 
subsequent work 
\cite{khashabi2020unifiedqa, paolini2021structured} \todo{Also cite https://arxiv.org/abs/1905.13453} and sequence labelling
showed that adapting T5-style models to particular domains results in powerful \mtl models. \citet{aghajanyan2020better} found that increasing the scale of a \mtl pre-finetuning step results in uniform improvement across various unseen tasks. 
Recent work in prompting large LMs has also shown that \mtl training can improve zero-shot performance \cite{wei2021finetuned, sanh2021multitask}. \citet{ye2021crossfit} showed that the few-shot performance on unseen tasks can be improved via a supervised \mtl step and recommended further analysis on task similarity and transferability. Our work aims to address this gap and connect 
large-scale \mtl learning to 
work on transferability.

\paragraph{Exploring Relationships Between Tasks}
\citet{wang-etal-2019-tell} and \citet{pruksachatkun2020intermediate} performed extensive empirical studies to identify the most beneficial intermediate tasks that improve target task performance both yielding mixed results. 
\citet{changpinyo2018multi} observed that jointly learning 11 sequence tagging tasks, with task embeddings, improves performance in around half of them and that the learned task embeddings revealed interesting task relationships such as clusters of semantic and syntactic tasks. \citet{kiperwasser2018scheduled} showed that learning to perform syntactic tasks such as dependency parsing and part-of-speech tagging along with translation in varying schedules improves translation performance---they used task embedding vectors to identify the tasks to the decoder model. Indeed the idea of using identifying tokens as language embeddings is known to improve translation \cite{johnson2017google} and dependency parsing \cite{ammar2016many} predates the widespread adoption of transformer models for these tasks.  

More recently, \citet{vu2020exploring} proposed two methods to learn task embeddings capable of predicting transferability between source and target tasks--one by pooling the representations of the textual task data from BERT and the other by using the layer-wise gradients of a BERT model. 
\citet{vu2021spot} 
learned task specific prompts that can benefit each other via prompt transfer. 
These works largely identify the single most suitable task for each target task, we extend the same to groups of tasks. 
Our work most closely relates to a concurrent study, \citet{aribandi2021ext5}, that examined the transfer across various task families. Our results compliment theirs using a different base model---they use a T5 style formulation of tasks, we use a shared Roberta encoder approach, showing that the transferability phenomenon is independent to the model architecture. We differ from their work in that we compare transfer on individual task groups with pairs of task groups as well and present results on the variance of performance as a result of \mtl learning.

%% file: conclusion.tex
\section{Conclusion}
In this work, we bring together the lines of exploration on transferability of tasks and large-scale \mtl training. 
Our results show that when the target tasks are unknown then \mtl scaling offers an effective way to obtain good downstream performance but if the goal is to improve performance on a specific target task set then a smaller set of related tasks is an effective, cheaper alternative.
We observe that task groups interact differently when combined and that selecting an optimum subset becomes harder as the size increases. 
We also see that variability across multiple random restarts decreases on related target tasks and also reduces on increasing the size of the \mtl step. 

%% file: appendix.tex
\section{Dataset Details}
\label{sec:datasets}
\paragraph{Sequence Labelling (SL) Tasks:} We source the following sequence labeling datasets and use them in the CONLL data format. \citet{ner-preprocessing} provides preprocessed data in the original splits. We report F1 score on all unseen target tasks. The total number of train examples across all 15 datasets is 225,433. \\

Pre-finetuning tasks:
\begin{itemize}
    \item CONLL - English, Spanish, Dutch, German \cite{sang2003introduction}
    \item Ontonotes - English, Chinese \cite{weischedel2013ontonotes}
    \item ANEM \cite{ohta2012open}
    \item GUM \cite{zeldes2017gum}
    \item GMB \cite{bos2017groningen}
    \item SEC Filings \cite{alvarado2015domain}
    \item Re3d  \cite{dstl-re3d}
    \item Malware TextDB \cite{lim2017malwaretextdb}
    \item Few-NERD \cite{ding2021few}
    \item SNIPS \cite{coucke2018snips}
    \item ATIS \cite{hemphill-etal-1990-atis} with preprocessing from the repository of \citet{hakkanitur2016multidomain}
\end{itemize}

Unseen Target Tasks:
\begin{itemize}
    \item WNUT17 \cite{derczynski2017results}
    \item Wikigold \cite{balasuriya2009named, nothman2013learning}
    \item MIT Movie Corpus - EN \cite{mitmovieen}
    \item NCBI Disease Corpus \cite{dougan2014ncbi}
\end{itemize}

\paragraph{Classification (C) Tasks} The pre-finetuning classifications tasks are from the Glue benchmark \cite{wang2019glue} and the 4 unseen tasks are from the Superglue benchmark \cite{wang2019superglue}. We use the versions made available via the Huggingface Datasets library \cite{lhoest-etal-2021-datasets, quentin_lhoest_2021_5639822}. In total, we have 943984 train examples across the pre-finetuning classification tasks. The unseen target tasks Superglue tasks, again made available via Huggingface. We retain the original splits and report performance on the validation sets. For Commitment Bank (CB) we report F1 score and accuracy for the other 3 unseen tasks as dictated by the metrics module from Huggingface. 

\paragraph{Question Answering (QA) Tasks} Our 6 pre-finetuning QA tasks \Cref{fig:tasks} are obtained from the MRQA dataset. We use the versions made available via the Huggingface Datasets library \cite{lhoest-etal-2021-datasets, quentin_lhoest_2021_5639822}. In total, we have 435,624 train examples across the pre-finetuning question answering tasks. We use the Books subset of the SubjQA dataset. For QED and SquadAdv, we split the data in a 4:1 ratio for train and validation. These datasets are collected as challenge sets explaining the big difference in performance. For XQuad, we use only the English language data and the original train and validation splits. We report F1 again as the metric of comparison for all QA tasks using the Huggingface metrics module.

The Huggingface Datasets library is released under the \href{https://github.com/huggingface/datasets/blob/master/LICENSE}{
Apache License 2.0}. The license information for all the sequence labelling datasets are available at \citet{ner-preprocessing}.

\paragraph{Heterogeneous Batches} When we run pre-finetuning on any combination of groups, we pool all the examples from the corresponding datasets and create hetergeneous batches from this pool. This is the chief reason for a gain in terms of computational cost on selecting a smaller subset and we provide statistics to measure this in \Cref{sec:cost}.

\section{Training Details} 

\subsection{Model Details}
\label{sec:app_model}
The model we use for our experiments is an XLM-Roberta model. We learn a shared encoder during pre-finetuning along with separate task specific heads for each pre-finetuning task. Our tasks are of three different formats that use the output of the encoder differently to make predictions---for classification we predict a single label for the entire sequence using the representation of the \textit{<s>} token, for sequence labelling we predict a label for each token in the input sequence and for extractive question answering we demarcate a span of the input sequence that corresponds to the answer for the question. The implementation of each of these follows from standard task-specific heads released by Huggingface where the corresponding output is fed to a linear classifier. When we initialize the model, we provide it with a list of pre-finetuning tasks to index the various task-specific heads. For each forward pass, the batch consists of the 
tokenized input 
as well as the task indices to be used for all the examples. When we use the model to report performance on a target task, the encoder checkpoint is loaded and a new task-specific head is initialized.

\subsection{Pre-Finetuning and Target Task Finetuning Details}
\label{sec:app_pft}

During pre-finetuning, the model sees examples from multiple tasks in the train and validation sets. We train each pre-finetuning model with an early stopping condition based on validation loss. We compute the validation loss for each task separately at the end of each epoch and 
the early stopping condition 
is when the average of the validation losses of all tasks does not improve for 3 epochs. 
This scheme is kept uniform for all pre-finetuning runs. We also keep the same batch size, 128, and search space of learning rates for pre-finetuning. We use validation data to search for the best learning rate, sweeping from $1e^{-3}$ to $1e^{-5}$. The rest of the Adam optimizer parameters are retained as the default values from Huggingface Trainer  \cite{wolf-etal-2020-transformers}. 

For target task finetuning, we load the saved pre-finetuned model and train the model to convergence defined as when the average validation loss does not improve for 3 consecutive epochs. We again sweep for the best learning rate from $1e^{-3}$ to $1e^{-5}$ and report performance on the best selection across 5 random restarts. 

\subsection{Computational Cost}
\label{sec:cost}
All experiments are run on an Amazon p3.16xlarge EC2 instance containing 8 Tesla V100 GPUs. The relative improvement in computational cost on selecting a smaller subset of tasks is mainly due to having fewer examples since we make use of hetergeneous batches. Following was the per-epoch runtimes during pre-finetuning on the various runs:
\begin{itemize}
    \item Only-SL - 1131 seconds per epoch
    \item Only-C - 1643 seconds per epoch
    \item Only-QA - 2237 seconds per epoch 
    \item SL+C - 2661 seconds per epoch
    \item QA+C - 3413 seconds per epoch
    \item SL+QA - 3079 seconds per epoch
    \item SL+QA+C - 4884 seconds per epoch
\end{itemize}

\section{Limitations and Future Work}
\label{sec:limitations}
\paragraph{Grouping of Tasks} We note that grouping of NLP tasks is often fuzzy and imprecise. Our chosen grouping was based on task format, as is used in recent work in the field \cite{vu2020exploring, sanh2021multitask}. We note that task similarity measures \cite{achille2019task2vec} calculate a more principled relationship between tasks.  We acknowledge that a more optimum grouping could be found for each target task set but our results (\Cref{tab:task_groups}) show that our chosen heuristic is a reasonable choice to isolate the effect of transferability. Selecting an optimal grouping would be a combinatorial task which would take significantly more compute. The way we choose to group tasks makes it easy to select 'related' tasks for each target task set. This might become more challenging for non-standard tasks.  

\paragraph{Controlling the Size of Each Task Group} Our chosen task groups have an unequal number of examples per task and group. We chose to retain all the examples from the various datasets since controlling for the number of examples also doesn't account for the relative difficulty of tasks (in particular we see that some of our chosen QA tasks seem to be more difficult for the model) but this could be a future line of research. 

\paragraph{Potential Risks} The main risks of our project are the risks associated with training large language models. We do not collect the datasets ourselves and use publicly released data which might contain biases against certain protected groups that will be reflected on models trained in this manner. To the best of our knowledge, these are standard datasets and we use them for the released tasks but we do not manually check them for offensive content.